\documentclass{ISMA_USD}

\hypersetup{pdftitle  = {On the topology and geometry of population-based SHM},
	pdfauthor = {K.\ Worden, T.A.\ Dardeno, A.J.\ Hughes \& G.\ Tsialiamanis},
	pdfkeywords = {ISMA2024, USD2024, PBSHM Paper}}

\newcommand{\ue}{\underline{e}}
\newcommand{\us}{\underline{s}}

\newcommand{\ux}{\underline{x}}

\newcommand{\uph}{\underline{\phi}}
\newcommand{\uth}{\underline{\theta}}

\newcommand{\C}{\mathbb{C}}

\newcommand{\R}{\mathbb{R}}

\newcommand{\sD}{\mathcal{D}}

\newcommand{\sT}{\mathcal{T}}
\newcommand{\sX}{\mathcal{X}}
\newcommand{\sY}{\mathcal{Y}}

\newtheorem{definition}{Definition}

\title{On the topology and geometry of population-based SHM}

\author{K.\ Worden, T.A.\ Dardeno, A.J.\ Hughes \& G.\ Tsialiamanis}

\affil{Dynamics Research Group, Department of Mechanical Engineering,\NewLineAffil
	   University of Sheffield, Mappin Street, Sheffield S1 3JD, UK}

\date{}

\graphicspath{{./Figures/}}
\begin{document}
%\bstctlcite{IEEEtran:BSTadapt}	

%%%%%%%%%%%%
% Abstract %
%%%%%%%%%%%%

% Enter your abstract in the command below.

\abstract{Population-Based Structural Health Monitoring (PBSHM), aims to leverage information across populations of structures in order to enhance diagnostics on those with sparse 
data. The discipline of {\em transfer learning} provides the mechanism for this capability. One recent paper in PBSHM proposed a geometrical view in which the structures were represented 
as graphs in a metric `base space’ with their data captured in the `total space’ of a vector bundle above the graph space. This view was more suggestive than mathematically rigorous, 
although it did allow certain useful arguments. One bar to more rigorous analysis was the absence of a meaningful topology on the graph space, and thus no useful notion of continuity. 
The current paper aims to address this problem, by moving to parametric families of structures in the base space, essentially changing points in the graph space to open balls. This allows 
the definition of open sets in the fibre space and thus allows continuous variation between fibres. The new ideas motivate a new geometrical mechanism for transfer learning in data are
transported from one fibre to an adjacent one; i.e., from one structure to another.}

\maketitle

\section{Introduction}
\label{sec:intro}

Although data-based {\em Structural Health Monitoring} (SHM) \cite{Farrar} is arguably established now as a means of embedding health assessment in the asset management process for 
engineering structures, there remain a number of implementation issues. One of the main barriers to widescale industrial uptake has proved to be the problem of obtaining training data 
which characterise the various health and damage states that an SHM system might be called upon to diagnose. One proposed means of overcoming the problem is to go from the monitoring 
of individual structures to populations; in this case, the probability that data are available for structural damage states for {\em some} structure in the population is potentially
greatly increased. If such data are available, {\em Population-Based SHM} (PBSHM) \cite{WordenP}, uses Transfer Learning (TL), to enable diagnostics on one 
structure, supported by data from another. 

\begin{figure}[htbp!]
    \centering
    \includegraphics[width=0.8\textwidth]{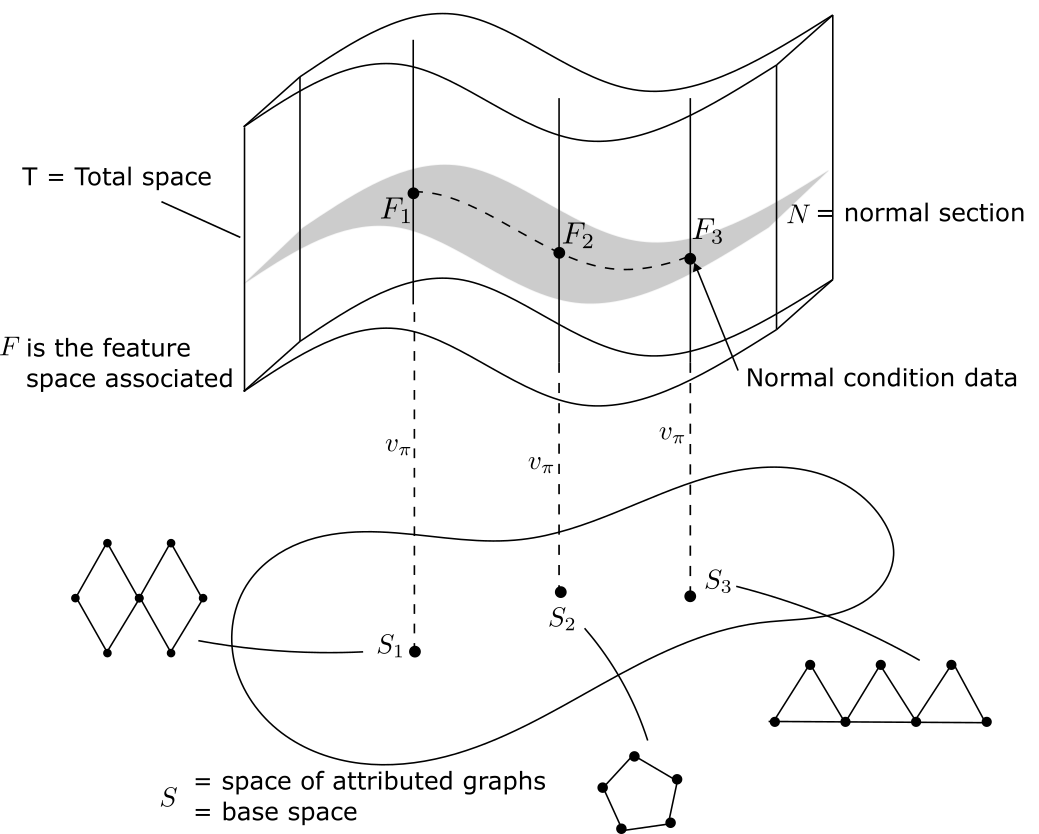}
    \caption{Schematic of fibre-bundle model of the PBSHM framework.}
    \label{fig:fb_scheme}
\end{figure}

The current paper is mainly concerned with the acquisition and use of data within the PBSHM framework; in particular it attempts to extend the geometrical picture of PBSHM presented 
in \cite{PBSHM4}; this picture is based on the idea of a {\em fibre bundle} from modern differential geometry \cite{Kobayashi}. In many ways, the fibre-bundle model is very natural 
for PBSHM as it can assign a separate feature space to each structure in a population; this is illustrated in Figure \ref{fig:fb_scheme}, where the `base space' -- which represents
the population -- is the set of structures of interest, with each represented as an attributed graph derived from an irreducible element model, as discussed in \cite{WordenP}; each 
structure $S_i$, is a single point in the base space. Data for a given structure are then considered to reside in the fibre $F_i$ above the point $S_i$, which will usually be a vector
space, and will usually be denoted $V$ from this point onwards. Movement of data for transfer learning between data spaces (as in domain adaptation \cite{PBSHM3}, for example), then 
corresponds to `horizontal' movement between fibres.

While the picture is suggestive and intuitive, it is of limited use as it stands, because it possesses little mathematical rigour. One major problem is that the base space of 
graphs is necessarily a point set with very little extra structure. Although the base space is a {\em metric space}, which is essential for PBSHM, it does not have a topology
and the picture thus lacks the important property of {\em continuity}. One of the objectives of the current study will be to extend the picture from \cite{PBSHM4}, so that 
continuous variation between structures is allowed; this in turn will allow movement between fibres, and thus a geometric interpretation of transfer. A second main objective of
this paper will be to use mathematical notation and language to explain how the bundle picture can be implemented in a practical framework which can embody PBSHM processes in
a natural manner. The idea here will be to present a theoretical framework for PBSHM which is backward-compatible with the studies so far and forward compatible so that it can 
guide development of the existing database formulation \cite{WordenP}.

\section{The geometry of feature spaces and data}
\label{sec:feat_data}

PBSHM is a data-based approach to SHM, so the first task in a systematic discussion must be to precisely describe the acquisition, storage and processing of data. Of course,
data will be assumed to come from {\em sensors}. The sensors would normally be assumed to be permanently installed on/in the structures of interest -- as is the norm in SHM
generally \cite{Farrar} -- or in a more generalised sense, consistent with the technology of non-destructive testing (NDT), could be thought of as instruments brought to the
structure. For now, sensors of the former type are assumed. 

Structures will be denoted by $S_i$, where $i$ indexes the structure within the population. Each structure will be represented within the population as an {\em attributed 
graph} (AG) derived from an {\em irreducible element} (IE) model, as discussed in \cite{WordenP}. In practice, a structure may have several representations within the population
corresponding to different IE models; however, for current purposes, this does not matter.

Each structure will usually have a set of associated sensors $M_i$, with individual sensors denoted $m_{ij}$; i.e., $M_i = \{m_{i1},\ldots,m_{i N_{M_i}} \}$, where $N_{M_i}$
is the number of sensors on the structure $S_i$. There is a danger of becoming overran by indices, so if the structure of interest is clear from context, the sensors can be 
denoted simply by $m_j$.

Each sensor will be assumed to deliver data; for the moment, these will be assumed to be records of sampled measurements. The main measurements will be samples of time data 
corresponding to state variables from the structure; these will be denoted $s_j(t)$, or $s_{jk}$ in discrete form\footnote{For consistency with other projects, the variable
notation is taken to be coincident with that from the study \cite{Worden}, which attempts to develop a theoretical framework for {\em digital twins}. At a fixed time $t$, the 
state variables can be assembled into a state vector $\us(t) = (s_1(t),\ldots,s_{N_{M_i}}(t))$. In the general SHM case, there will be some sensors which measure environmental 
states which can be used to compensate for environmental and operational variations \cite{Farrar}; these variables are denoted $e_i(t)$ and can be assembled into a 
corresponding environmental state {\em vector} $\ue(t)$.}, where $k = 1,\ldots,N_{T_j}$ and $N_{T_j}$ is the number of samples in the record for $s_j$. For now, it is sufficient
to say that the sensors will record sample sets intermittently or periodically, but will always save the same number of points $N_T$. This is the usual practical choice; for 
example, in the classic Z24 Bridge-monitoring case \cite{Peeters}, the sensors collected records at 10-minute intervals. The measurement system recorded fixed-time sample 
records from a set of accelerometers, which were then used for modal analysis using the stochastic-subspace identification (SSI) algorithm. As `feature extraction' methods like
SSI are more easily applied in the situation, it will be assumed here that the time-series data satisfy three conditions:

(i) Each sensor record has the same length: $N_{T_i} = N_{T_j} = N_T$.

(ii) Each record is sampled with the same frequency; $f_s = 1/\Delta t$.

(iii) For a given structure, each record is initially synchronised: $t_{i1} = t_{j1}$ for sensors $m_i$ and $m_j$.

Clearly, conditions (ii) and (iii) together ensure that records remain synchronised.

Time data of this nature will be referred to here as `raw', `base' or `initial' data.

\begin{figure}[htbp!]
    \centering
    \includegraphics[width=0.6\textwidth]{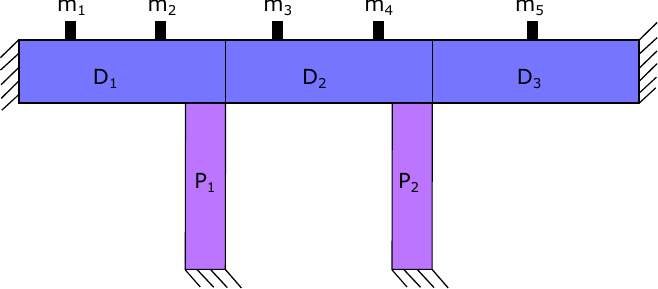}
    \caption{Cartoon representation of a three-span bridge IE model with sensors. Green circles are ground nodes.}
    \label{fig:3span_ie}
\end{figure}

\begin{figure}[htbp!]
    \centering
    \includegraphics[width=0.6\textwidth]{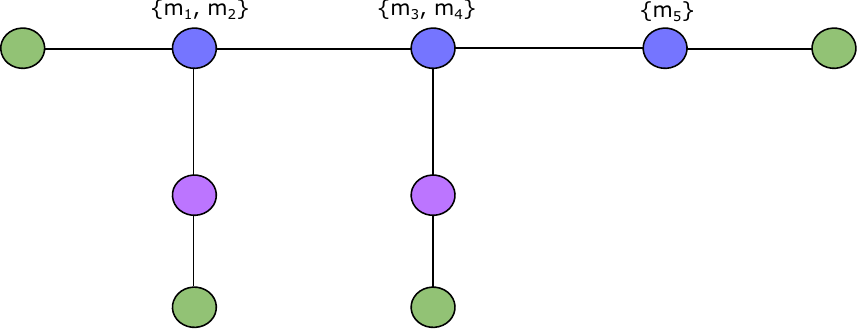}
    \caption{Attributed graph of three-span bridge IE model.}
    \label{fig:3span_ag}
\end{figure}

Later, when sensor placement optimisation is considered, it will be important to know where sensors are located. For the purposes of forward compatibility, it will be assumed
that sensors can be located to within an IE. For now, it will be assumed that sensors are {\em always} associated with an IE, and can be located within it using the vector
of attributes. This condition means that sensors are always associated with vertices in the AG, and never edges. This restriction can be removed if necessary at a later date.
As an illustration, consider the three-span bridge model in Figure \ref{fig:3span_ie}, which will have an AG as shown in Figure \ref{fig:3span_ag}.

The discussion can now move on to the data spaces. Because a data record from a sensor will comprise $N_T$ measurements of real numbers, each record can be considered as an 
element in a vector space $\R^{N_T}$. An individual state-variable record from sensor $m_j$ will be denoted by $\us_j \in \R^{N_T}$. Over time, the database will accumulate $N_R$
records from each sensor and the total saved data can be represented as a matrix $S_{jk} \in \R^{N_t} \times \R^{N_R}$, where $j$ labels the sensor and $k$ labels the record.
The index $j$ will also be referred to as a {\em channel}. $k$ also labels the {\em time} of the measurement activation $\tau_k$; it will be assumed here -- as in the Z24 case --
that activations are periodic, so $\Delta \tau = \tau_k - \tau_{k-1}$ is constant, and $\Delta \tau >> \Delta t$.

The population case will require more definitions and terminology, but for now the discussion of data will concentrate on the individual structure case. 

In terms of the total data space for a given structure at a time $\tau$, the {\em raw} data are records of time data $\us = \{s_1,\ldots,s_{N_T}\}$; these are the records which
will be subject to any subsequent signal processing to derive {\em features} for SHM diagnostics. These data are points in a vector space $V_S \cong \R^{N_T}$. Now, if there are 
$N_S$ sensors on the structure and these are all triggered $k$ times, the total data space for a structure $S$ is given by,

\begin{equation}
     V_S = \bigoplus_{j,k} V_{T_{jk}}
\label{eq:tds}
\end{equation}
which is the direct sum of $jk$ copies of $V_T$. This is potentially a space of very high dimension, as $\mbox{dim}(V_S) = j k N_T$. This picture summarises the data for a given
structure at the current acquisition point, and represents the vector space of data that lives `above' a given structure in the bundle picture. These data will be involved in
the machine-learning operations used to classify health states for SHM. The record index $k$ is important for machine learning as it can be used to partition the data into
training, validation and testing sets. At the risk of more indices, a given time series for processing can be denoted by $\us^{jk} = \{s_1^{jk},\ldots,s_{N_t}^{jk}\}$; i.e.,
the time data associated with sensor/channel $j$ at acquisition time $k$.

To recover specific time series from the body of data, one can define {\em projection operators} $\pi_j$ and $\pi_k$, where,

\begin{equation}
     \pi_j : V_S \longrightarrow \bigoplus_k V_{T_{jk}}
\label{eq:defpij}
\end{equation}
selects the records corresponding to channel $j$, and,

\begin{equation}
     \pi_k : V_S \longrightarrow \bigoplus_j V_{T_{jk}}
\label{eq:defpik}
\end{equation}
selects all the channels at acquisition time $k$. Note that $\pi_j$ and $\pi_k$ both act on $V_s$, so they cannot be combined as $\pi_j \circ \pi_m$. To extract a specific
channel at a specific time, one needs a specific $\pi_{jk} : V_S \longrightarrow V_{T_{jk}}$, then $\pi_j(V_S) = \oplus_k \pi_{jk} (V_S)$ etc.

The reader might argue that this looks like a lot of symbolism just for the sake of it, and there might some justice in that; however, the PBSHM software framework is going
to need to select specified data records for signal processing, and the projection operators defined here correspond to selection functions, and thus provide a bridge between
the mathematics of PBSHM and the software. 

\begin{figure}[htbp!]
    \centering
    \includegraphics[width=0.6\textwidth]{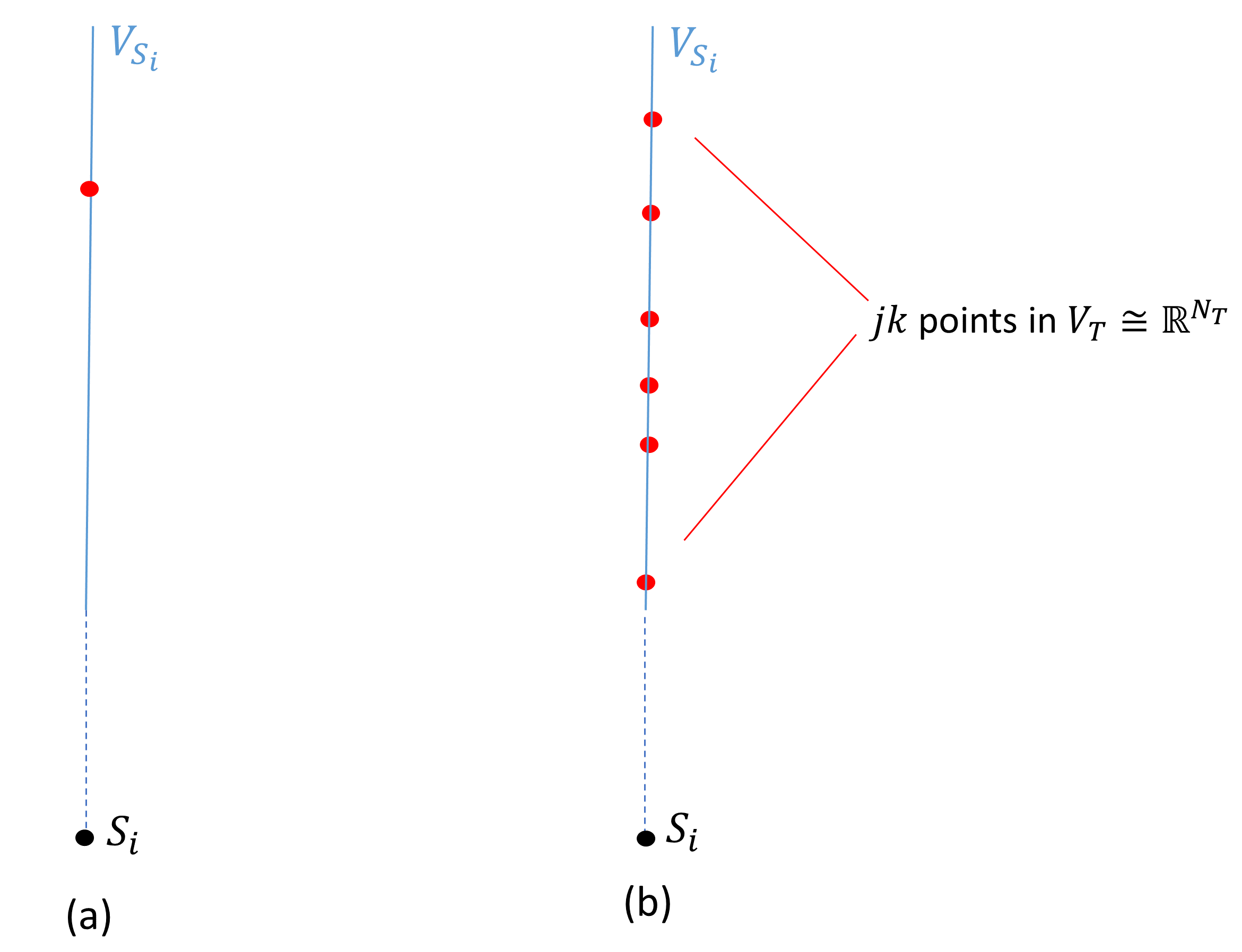}
    \caption{Schematic for two ways of thinking about the data fibres.}
    \label{fig:datafibres}
\end{figure}

At this point, one can think of the fibres and data in two useful ways, as shown in Figure \ref{fig:datafibres}. In the first picture, as in Figure \ref{fig:datafibres}a, one 
can record the totality of data for the structure $S_i$ as represented by a single point in a very high-dimensional fibre $V_{S_i}$ with dimension $N_S \times N_T \times N_R$.
In the second picture, there is a point for each actual measurement, so $jk$ points and $\mbox{dim}(V_{S_i}) = N_T$. The first picture is useful in terms of book-keeping, while the
second corresponds to the usual viewpoint in machine learning for visualisation of data sets. Moving between these pictures uses the projection operator $\pi_{jk}$. 

The overall viewpoint now is of the raw data, before any transformation into damage-sensitive features, so the question arises of how to label and store derived feature data.
Feature vectors will be the result of applying signal-processing operations to the raw time-series data $\us_{jk} \in \R^{N_T}$; i.e., in the second picture of the data fibres.
In the PBSHM framework, it is assumed that the operations are drawn from a labelled set of $N_O$ possible options $\{O_1,\ldots,O_{N_O}\}$. For example:

(i) $O_1$ extracts the mean of the signal, $O_1(s_{jk}) = \overline{s}_{jk}$. \\
(ii) $O_2$ extracts the frequency spectrum using the Discrete Fourier Transform (DFT). \\
(iii) $O_3$ extracts the spectrum via a Welch algorithm with window-length $N_w$.

Each operator is a map $O_i : \R^{N_T} \longrightarrow \R^{N_{O_i}}$, so:

$O_1: \R^{N_T} \longrightarrow \R$ \\
$O_2: \R^{N_T} \longrightarrow \R^{N_T} \cong \C^{N_T/2}$ \\
$O_3: \R^{N_T} \longrightarrow \C^{N_w}$

\begin{figure}[htbp!]
    \centering
    \includegraphics[width=0.6\textwidth]{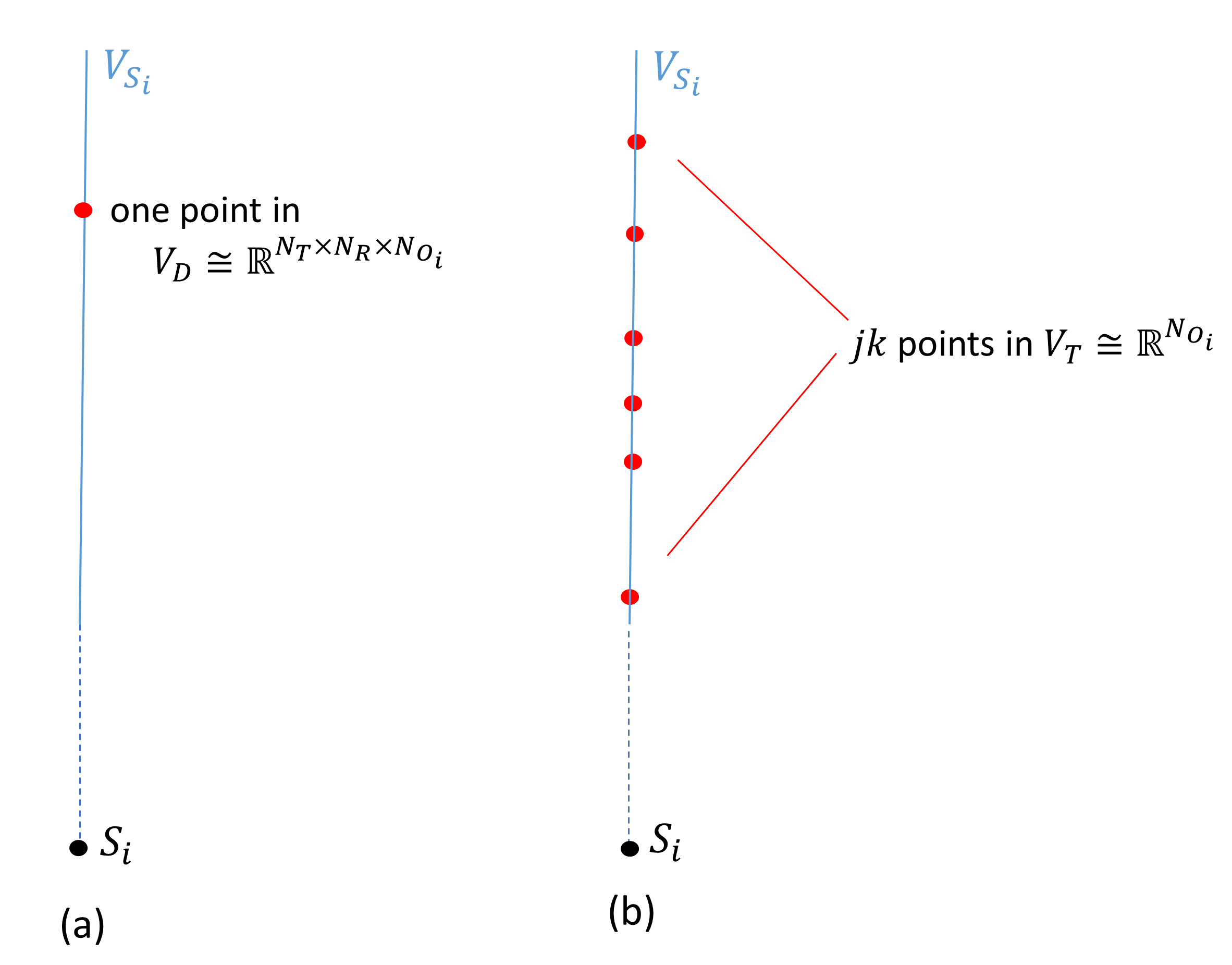}
    \caption{Schematic for two ways of thinking about the feature data fibres.}
    \label{fig:featfibres}
\end{figure}

As before, there will be two possible pictures for feature data in terms of data fibres (Figure \ref{fig:featfibres}).

Now, it will be clear that one might want to carry out a sequence of signal-processing procedures over time. In the spirit of traceability as exhibited by the Los Alamos ECHO dataspace,
the idea will be that {\em all} feature spaces and their intermediate states will saved in the PBSHM database framework; this allows that all data can be potentially recovered later and
used for transfer. The original `raw' data and all derivatives are stored and labelled; this gives a `stratified' data structure for each actual structure of interest. As above, the 
raw data space will be labelled $V^{(0)}_{S_i}$; derived feature spaces will then follow the operations of feature extraction, denoted $O^{(i)}$ in sequence; i.e., the first derived 
space would be denoted $V^{(1)}_{S_i} = O^{(1)} V^{(0)}_{S_i}$, in an obvious notation. Note that not all derived features need to be used for machine learning, some will represent
intermediate operations like mean removal. These are saved for transparency/traceability, and also because mean-removal might be the first step in a different feature extraction
operation and the data can be reused. The total data space, including derived features will thus be,

\begin{equation}
     V^*_{S_i} = \bigoplus_{m=0}^M V^{(m)}_{S_i}
\label{eq:tdsp}
\end{equation}

Data required for ML will be recovered via another projection operator $\pi^{(j)} : V^*_{S_i} \longrightarrow V^{(j)}_{S_i}$. This operation, followed by $\pi_{jk}$ generates 
specific feature data for machine learning, as illustrated in Figure \ref{fig:featfibres}b.

This is all rigorous so far -- if somewhat mathematically vacuous -- and captures the data spaces over structures and projection operators which will select features for signal
processing and transfer learning; the discussion can proceed to transfer itself.

\section{Transfer between data spaces}
\label{sec:transfer}

In the fibre-bundle picture, the population is encoded in a complex network, which is essential a large graph in which individual structures are the nodes (which are themselves
attributed graphs). Above each node is a fibre which contains the entire data inventory for that structure. The next step is to describe how to implement transfer learning.

Two objects are required to formally define transfer learning (TL) \cite{PanB},
	
\begin{itemize}
\item[$\bullet$] A {\bf domain} $\sD = \{\sX,p(X)\}$, consists of a feature space $\sX$ and a marginal probability distribution $p(X)$ over the
                 feature data $X=\{\ux_i\}_{i=1}^N \in \sX$, a finite sample from $\sX$.
\item[$\bullet$] A {\bf task} $\sT=\{\sY,f(\cdot)\}$, consists of a label space $\sY$ and a predictive function $f(\cdot)$\footnote{Clearly, $f(\cdot)$ is the core
of the learning problem under consideration. In a regression problem this might simply be some desired map from $\sX$ to $\sY$, where $\sY$ is a continuous label
space. In the context of a classification problem, the prediction function of interest will often be the conditional probability $p(y|\ux)$.} which can be inferred
                 from training data $\{\ux_i,y_i\}_{i=1}^N$ where $\ux_i \in \sX$ and $y_i \in \sY$.
\end{itemize}
	
\noindent Using these objects, transfer learning between a task in a single source domain and another in a single target domain is defined as \cite{Pan},
	
\begin{definition}{{\bf Transfer learning}}
is the process of improving the target prediction function $f(\cdot)$ in the target task $\sT_t$ using knowledge from a source domain $\sD_s$ and a source
task $\sT_s$ (and a target domain $\sD_t$), whilst assuming $\sD_s \neq \sD_t$ and/or $T_s \neq \sT_t$.
\end{definition}

Clearly, a problem will occur if the training data for an algorithm have a different feature distribution to the test data; for example, any decision boundaries
established during training may be seriously in error during testing. This particular type of discrepancy between problems is referred to as {\em domain shift}.
In the context of PBSHM, one might well expect domain shift between the feature spaces of the source and target structures.

\noindent Within the field of transfer learning, {\em domain adaptation} arguably offers one of the most useful tools for PBSHM and is defined as \cite{Pan2},
	
\begin{definition}{{\bf Domain adaptation}}
is the process of improving the target prediction function $f(\cdot)$ in the target task $\sT_t$ using knowledge from a source domain $\sD_s$ and a source
task $T_s$ (and a target domain $\sD_t$), whilst assuming $\sX_s = \sX_t$ and $\sY_s = \sY_t$, but that
$p(X_s) \neq p(X_t)$ and typically that $p(Y_s\;\vert\;X_s) \neq p(Y_t\;\vert\;X_t)$.
\end{definition}

\begin{figure}[htbp!]
    \centering
    \includegraphics[width=0.4\textwidth]{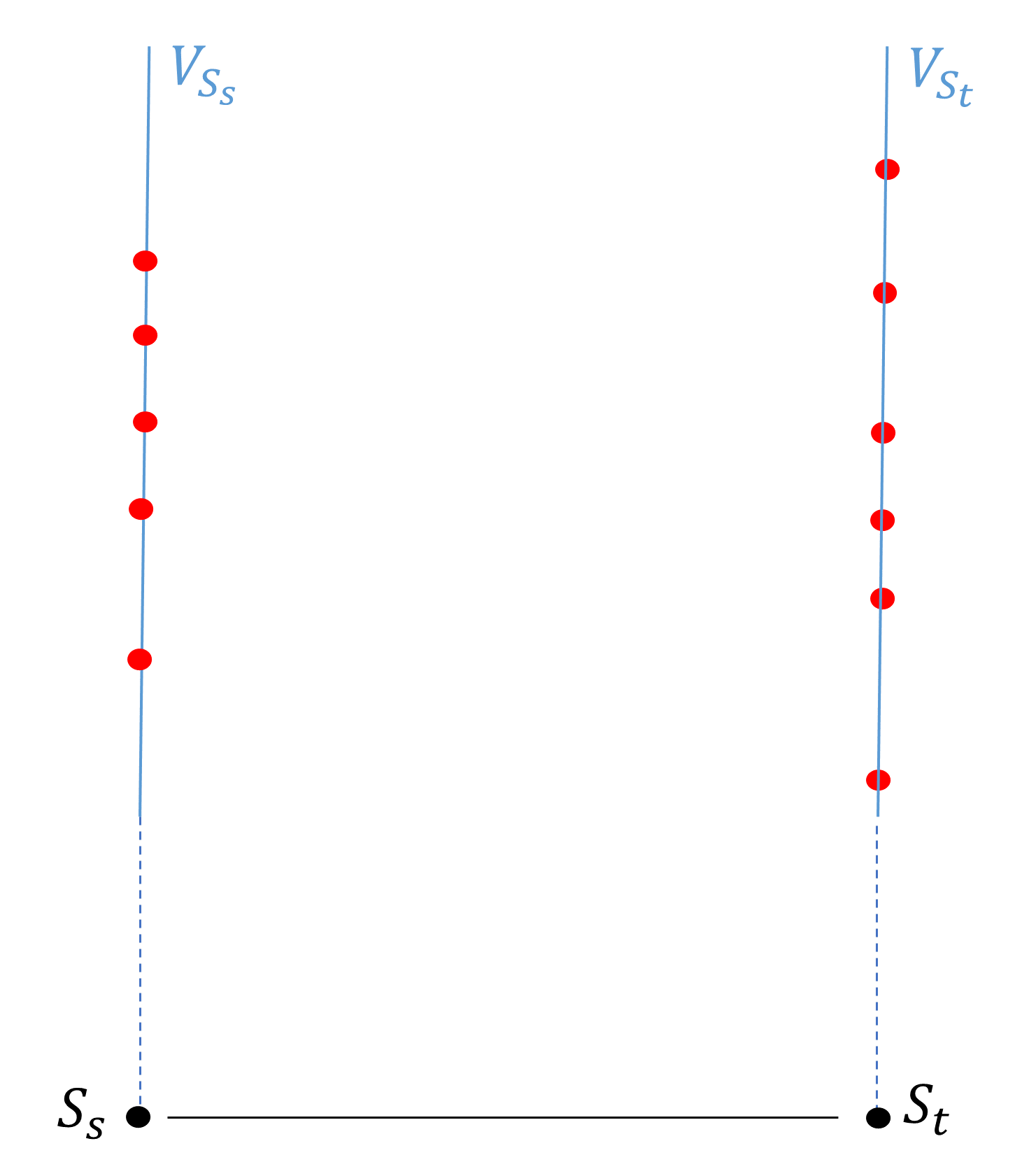}
    \caption{Schematic of data fibres for transfer between two structures.}
    \label{fig:ssst}
\end{figure}

Given these definitions, it would appear that it is sufficient to consider two structures from the population at the moment; a source and a target, as illustrated in 
Figure \ref{fig:ssst} with their data spaces/fibres.

To simplify matters, it will be assumed that the data for the structures $S_s$ and $S_t$ will have the same label spaces, but the data for $S_t$ will not include labels. The
TL problem here is then to train a classifier on $S_s$ data that will work on the $S_t$ data. There are essentially two ways to do this:

\begin{enumerate}
\item Map the $S_t$ feature data into the $S_s$ feature space via some constructed $\phi$ and hope that the decision boundaries there will work for $S_t$.

\item Map both sets of data via a $\phi$ into a common latent space $V^*$, in which a classifier trained on the $\phi(S_s)$ will work on the $\phi(S_t)$ data.
\end{enumerate}

\begin{figure}[htbp!]
    \centering
    \includegraphics[width=0.4\textwidth]{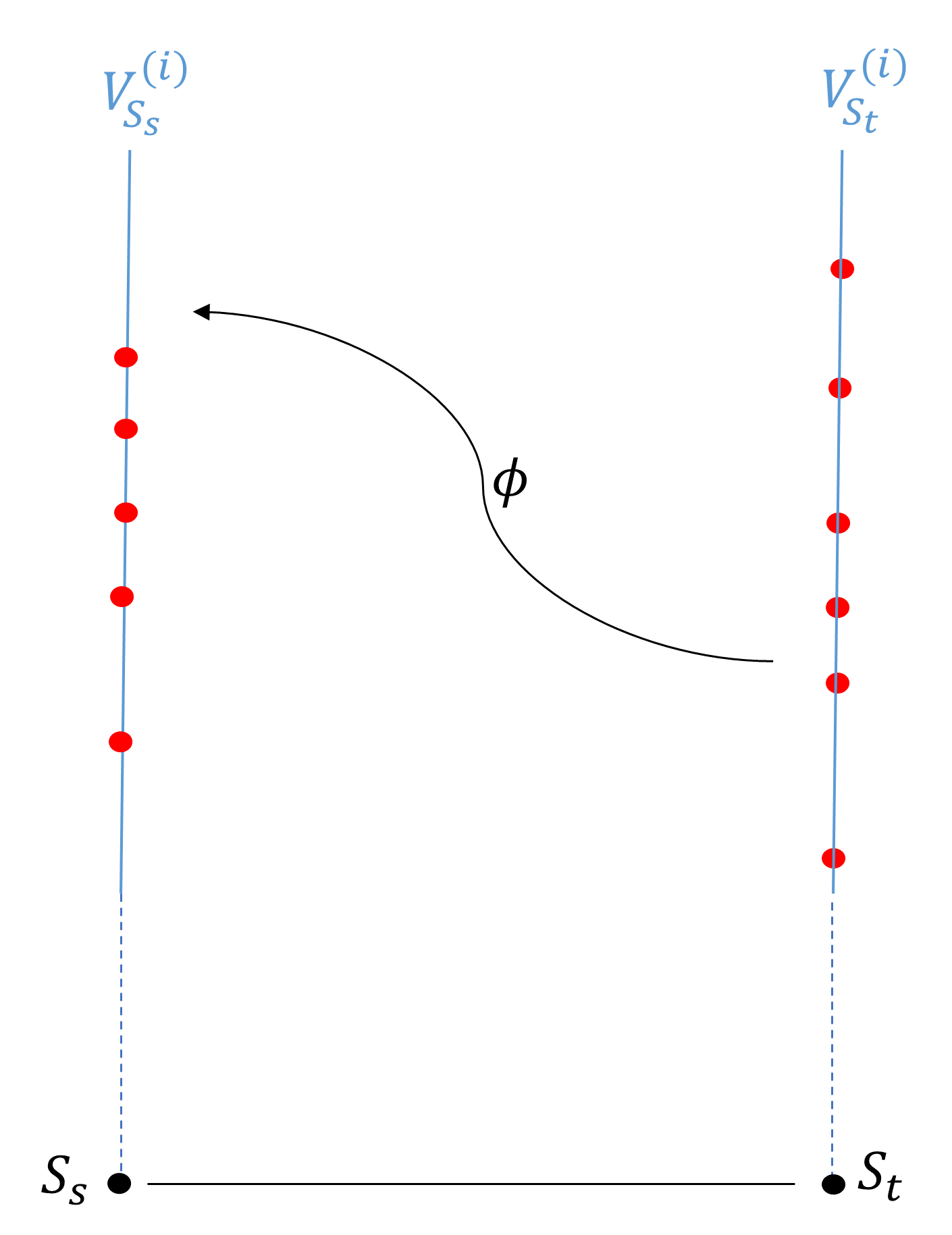}
    \hspace*{10mm}
    \includegraphics[width=0.4\textwidth]{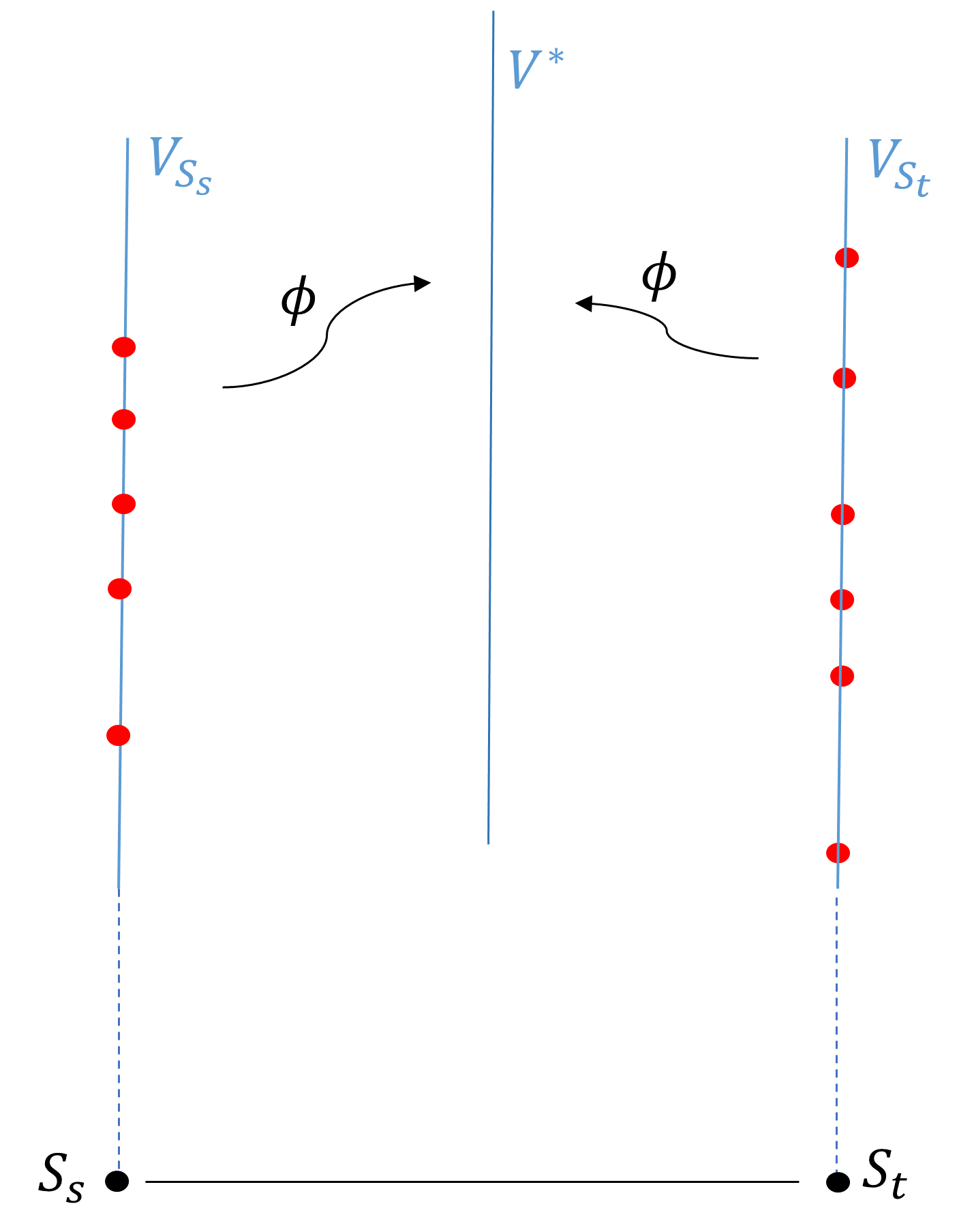}
    \caption{Transfer mechanisms: (a) Direct domain transfer; (b) Domain adaptation.}
    \label{fig:ddt}
\end{figure}

Option 1 (Figure \ref{fig:ddt}) will be referred to here as {\em direct domain transfer} (DDT), while Option 2 is clearly the standard domain adaptation (DA) referred to above. 

At this point, the discussion does not appear to involve geometry or any real geometric PBSHM; however, one geometrical concept has been mentioned. The space of structures is 
assumed to be a {\em metric} space, in which one can measure the `distance' between or similarity of structures. This detail is important as it will be assumed that transfer 
will be more likely to be successful between two structures which are `close' in the metric. The metric will be denoted by $D(S,S')$ here; there are various possible metrics,
the one discussed in \cite{WordenP} is based on the maximum common subgraph between $S$ and $S'$. This idea motivates a rationale for choosing source structures from the 
population.

Given a target structure $S_t$, which is data-poor, and a candidate population of data-rich structures $S_i$, one might select the best source for transfer, via

\begin{equation}
     S_s = \underset{i}{\mbox{argmin}}~ D(S_i,S_t)
\label{eq:min_dist}
\end{equation}

One then needs a threshold $d_s$ such that one only expects transfer to be successful with given probability if $D(S_s,S_t) \ge d_s$.

This measure of similarity is a good use of geometry; however, one might ask if more is possible. However, at the moment the fibres are not linked together and DA etc.\ do not
use geometry. The problem is that there is no real geometry to use. The base space has no usable topology, and thus no obvious notion of continuity; the total space is basically
the disjoint union of a number of vector spaces. If there were some way of mapping between the fibres, one would have geometrically-motivated DDT. The question is how to add
a more usual bundle structure to the space of structures.

The first issue is that all the fibres would need to be homeomorphic to some standard fibre and thus have the same dimension. This is not a major problem because, intuitively, 
one would expect the best results to follow when the source and target feature spaces are {\em compatible}, in that they represent the {\em same physics}; i.e., one might wish to 
look at the first four natural frequencies on both structures, or some number of spectral lines from their FRFs. The strategy relies on a straightforward use of the projection
operators. Suppose that the metric has singled out $S_s$ as the nearest possible source structure. Furthermore, suppose that the relevant data spaces/fibres are then
$V_{S_t} = \oplus_m V_{S_t}^{(m)}$ and $V_{S_s} = \oplus_n V_{S_s}^{(n)}$; in this case, one projects out pairs of feature spaces that have undergone the same chain of 
signal-processing operations; i.e., $O^{(i)} = \prod O_i = \{\mbox{extract the first four natural frequencies}\}$. In this case, the transfer problem is as shown in Figure
\ref{fig:ddt}, and the fibres have the same dimension, as required for the structures taking part in the TL.

Of course, what would be very useful would be some help from the geometry in getting the mapping $\phi$. The problem, as mentioned earlier is that the complex network that forms
the base space has no usable topology as defined - no notion of continuity. In a normal vector bundle \cite{Kobayashi}, one could define a {\em connection} to get between fibres.
Roughly speaking, the tangent space of the total space could be divided locally in to a {\em vertical subspace} $V$ and and orthogonal {\em horizontal subspace} $H$, where the 
tangent directions on $V$ would point along the fibre and the directions on $H$ would point {\em between the fibres}. The problem here is that $H$ just inherits the discrete
topology from the base space and doesn't have a notion of continuity either. Ideally, one would like to transfers between structures that are both points in something that looks
locally like a manifold; how might one do this?

\begin{figure}[htbp!]
    \centering
    \includegraphics[width=0.5\textwidth]{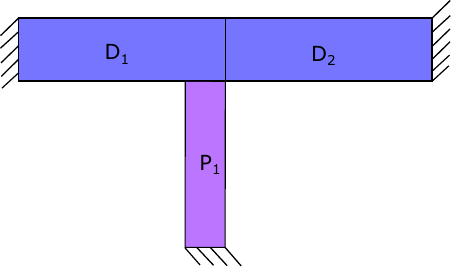}
    \caption{'Cartoon' representation of a two-span bridge $B_2$.}
    \label{fig:two_span}
\end{figure}

The solution is to regard structures as belonging to {\em parametric families}. To illustrate this idea, consider Figure \ref{fig:two_span}, which represents a `cartoon' 
IE model of a two-span bridge $B_2$. 

Suppose for simplicity that each irreducible element in the model is a cuboidal slab or pillar; each will have dimensions length, width and thickness $\{l,w,t\}$ and a minimal
set of material properties $\{E,\rho,\nu\}$. The total two-span structure will thus be parametrised by a real vector,

\begin{equation}
     \uth = (l_1, w_1, t_1, E_1, \rho_1, \nu_1, l_2, w_2, t_2, E_2, \rho_2, \nu_2, l_3, w_3, t_3, E_3, \rho_3, \nu_3)
\label{eq:theta_2sb}
\end{equation}

If one bridge is allowed for each possible $\uth$, the point specified by a graph in the space of structures is expanded into an entire open 18-dimensional hypercube of two-span
bridges, where each parameter takes values from some open interval of allowed values. A simple FE model -- for example -- can then be constructed for the IE model in order to 
provide feature values with which to populate the fibres above the points $\theta$; i.e., to solve for the first four natural frequencies as a function of the material properties
and dimensions. Now one has something which looks (at least locally) like a {\em bona fide} vector bundle.

\begin{figure}[htbp!]
    \centering
    \includegraphics[width=0.6\textwidth]{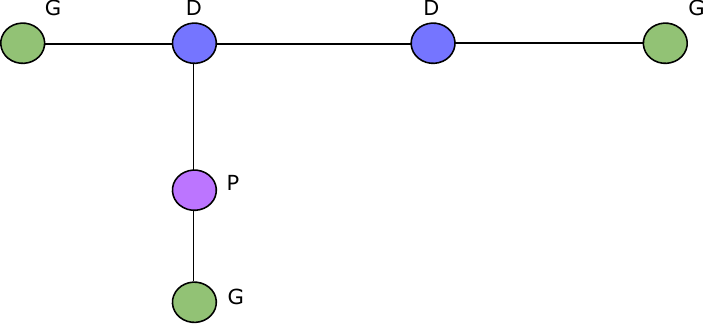}
    \caption{Attributed Graph (AG) representation of a two-span bridge $B_2$, where `G' represents a ground node, `D' a deck node and `P' a pillar.}
    \label{fig:two_span_ag}
\end{figure}

Now, if the SHM problem of interest involved a cartoon two-span bridge as the target structure, one may well believe that the candidate source structures would also be two-span
bridges and would have the same topologies as AG representations (Figure \ref{fig:two_span_ag}). In this case, the source and target would both be over the same parametric family and 
would thus be joined by a continuous path; for all intents and purposes, the transfer problem would be specified in terms of a standard vector bundle.

\begin{figure}[htbp!]
    \centering
    \includegraphics[width=0.4\textwidth]{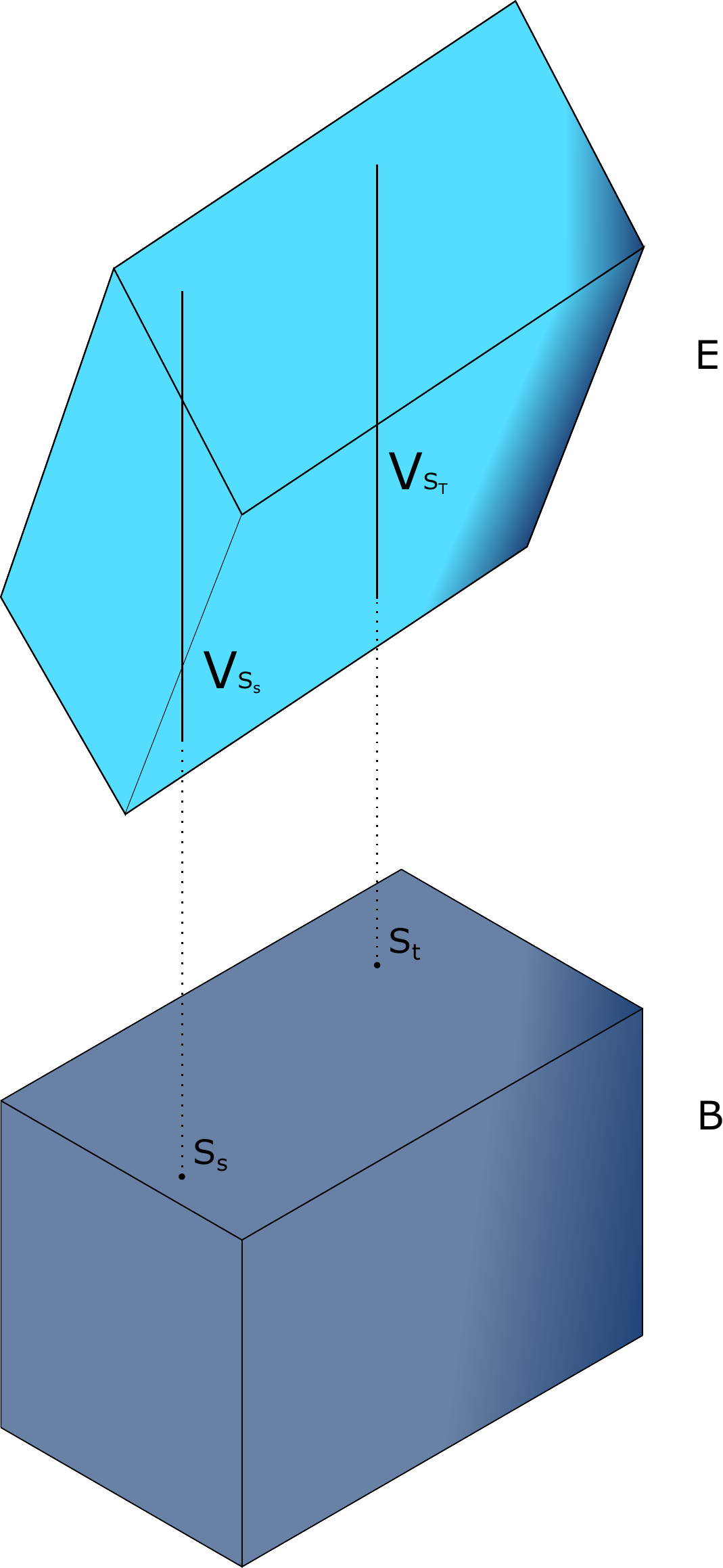}
    \caption{Vector bundle of feature spaces over a population formed from a parametric family of structures.}
    \label{fig:vb_pf}
\end{figure}

On structural topology alone, the distance between $S_s$ and $S_t$ would be zero; adding the attributes to the metric (using a simple Euclidean norm) would mean that the closest 
source structure would have the closest dimensions and material properties; one would have the situation illustrated in Figure \ref{fig:vb_pf}. In this case, a true vector 
bundle is obtained, with both the base $B$ and total $E$ spaces manifolds with boundary -- in this case hypercubes.

As movement between the structures and the fibres is all taking place within open sets and locally in an open ball around both $S_s$ and $S_t$, one can potentially use the 
geometry to construct fibre maps for DDT; discussion on how this might be accomplished is postponed for later work.

\begin{figure}[htbp!]
    \centering
    \includegraphics[width=0.6\textwidth]{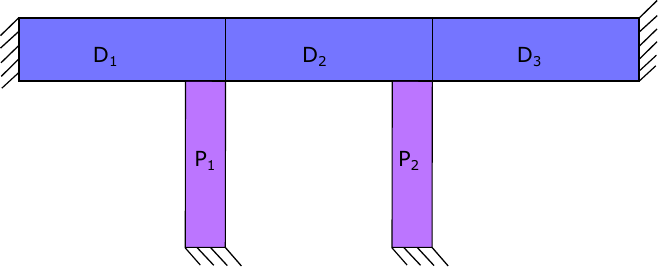}
    \caption{IE-model representation of a three-span bridge $B_3$ with three deck elements and two pillars.}
    \label{fig:three_span_ie}
\end{figure}

Now, suppose that one is faced with a real problem involving a two-span bridge as the target structure, but the database does not contain any data for such a structure. One
might imagine that the `closest' object in the population is a three-span bridge $B_3$ with an IE model as in Figure \ref{fig:three_span_ie}.

\begin{figure}[htbp!]
    \centering
    \includegraphics[width=0.45\textwidth]{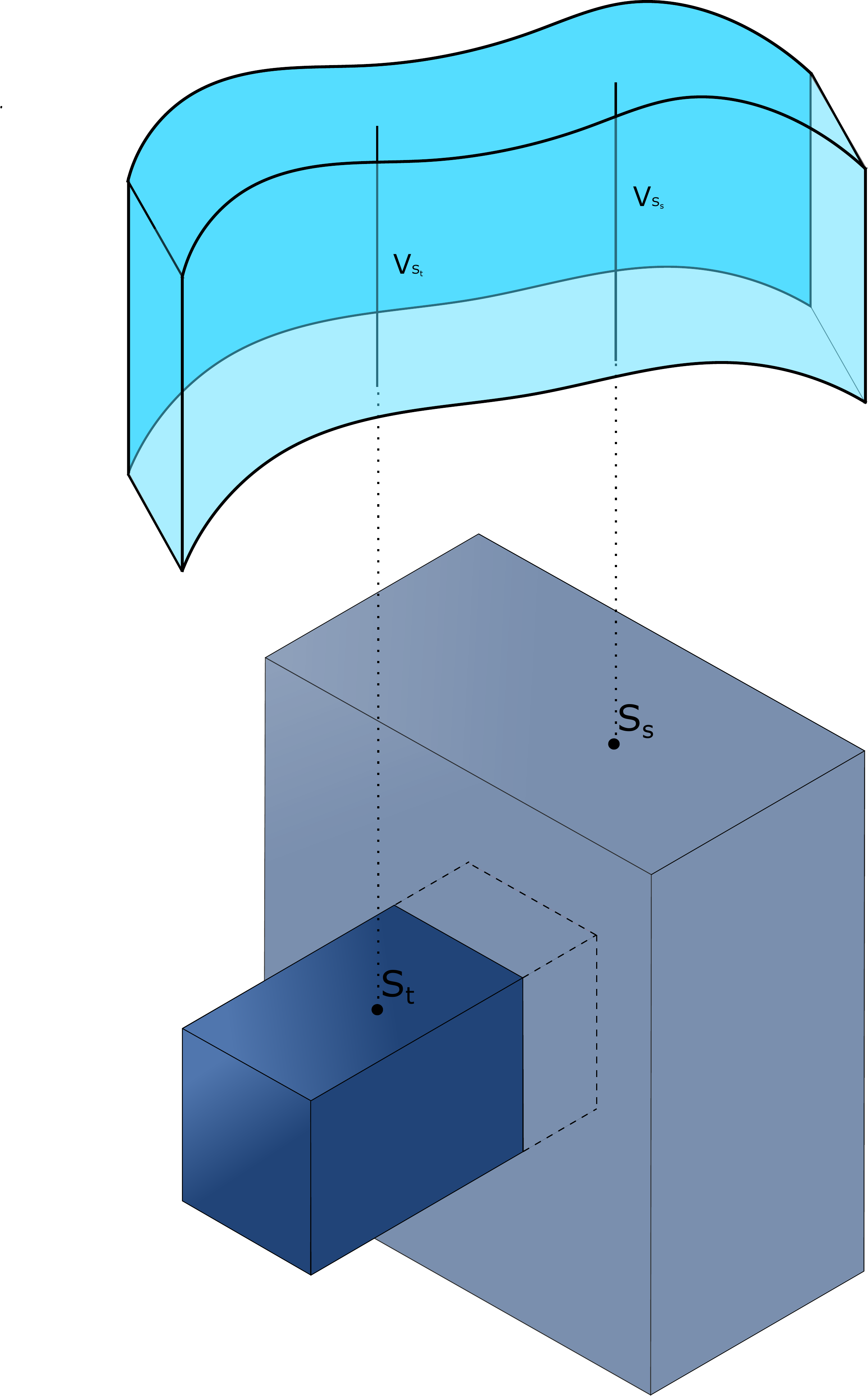}
    \caption{Geometry of transfer problem between a two-span and a three-span bridge.}
    \label{fig:tf_23}
\end{figure}

If one assumes the same classes of geometrical and material parameters for the three-space bridge, one can see that it belongs to a parametric family $\uph \in \R^{30}$.
The situation for transfer is now as shown in Figure \ref{fig:tf_23}. As before, one would demand consistent features, so the dimensions of the fibres would be the same over
both hypercubes, even though the points in the base space would have different dimensions.

Now, note that one can actually continuously deform bridge $B_3$ into $B_2$, by allowing the parameters $l_{P2}$ and $l_{D3}$ (in an obvious notation) to tend to zero. One 
could also allow P2 to be removed by allowing its material properties to go to zero; however, this would be less physical. Furthermore, one cannot simply let the material properties
of deck element D3 to tend to zero, as this would disconnect element D2 from the ground. 

\begin{figure}[htbp!]
    \centering
    \includegraphics[width=0.4\textwidth]{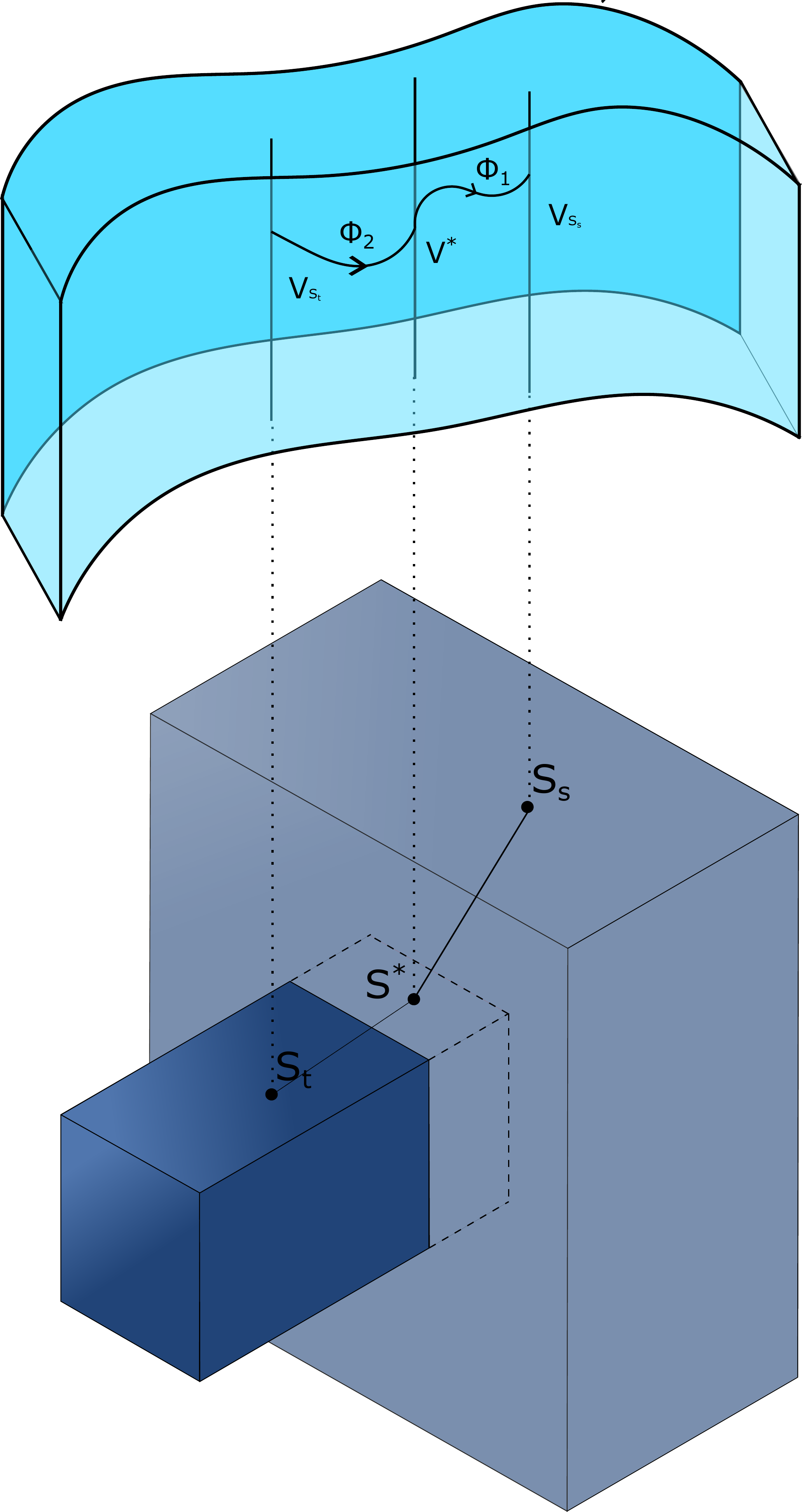}
    \caption{Transfer problem between a two-span and a three-span bridge, showing continuous deformation between problems.}
    \label{fig:tf2_23}
\end{figure}

The crucial point here is that, within the base space, the situation is as shown in Figure \ref{fig:tf2_23}, where $B_3$ can be continuously deformed into a bridge that is in
the same parametric family as $B_2$, denoted here as $B_2^*$. The important thing to note is that, because the fibres (features) are consistent, one can construct a map
$\phi_2: V_{S_t} \longrightarrow V^*$ over the parametric family which contains $B_3$, and then another $\phi_1: V^* \longrightarrow V_{S_s}$ which together allow DDT via
the composition $\phi_1 \circ \phi_2 : V_{S_t} \longrightarrow V_{S_s}$.

In fact, if one wished to set a maximum dimension for the parametric families, one could embed the lower-dimensional families into the higher by simply setting some groups
of parameters to zero for different classes of structures. This would have the marked advantage of making the space of structures into a standard single vector space of
constant (high) dimension. The disadvantage of this approach would be in doing the book-keeping between radically-different classes. While the approach might be straightforward
for a population of multi-span bridges, it would likely be challenging if the population were very heterogeneous; i.e., involving both bridges and aircraft.

The constructions here raise the possibility of an interesting idea -- that of {\em interpolating structures}.

\section{Interpolating structures}
\label{sec:interpol}

Consider a situation where one has a new target structure $S_t$ to monitor, with little or no data, so that transfer is indicated. Further suppose that previous analysis has
established a threshold distance $d_s$ for positive transfer, but that {\em no} structures in the population are within the required distance; i.e., there is no 
candidate source $S_s$ such that $D(S_s,S_t) \le d_s$. 

One can conceive now of {\em creating} a {\em computational model} structure $S^*$ such that $D(S_s,S^*) \le d_s$ and $D(S^*,S_t) \le d_s$; both individual transfers are now 
potentially valid and the required transfer might be accomplished in two steps. This idea is feasible because the PBSHM Framework {\em does not} distinguish between real and model 
structures. Furthermore, if $S^*$ were specified as an FE model, for example, one could potentially generate as much feature data as were needed by simulation.

The design of the required $S^*$ is likely to require some insight into PBSHM problems. One obvious strategy would be to choose $S_s$ as the closest structure to $S_t$
even if $D(S_s,S_t) > d_s$ and then design an intermediate or interpolating structure. Where a clear way forward exists is if the parametric family for $S_t$ is 
{\em contractible} to the family for $S_s$ by allowing some subset of parameters to go to zero and others to move. This places $S^*$ in the overlap of both families as 
discussed earlier and can potentially be carried out in such a way that brings $S^*$ closest to $S_s$, in the awareness that the object of interest here is really
$D(S_s,S^*) + D(S^*,S_t)$. 

One can profitably think about the {\em paths} between structures now. Suppose that some parameters $\uth^0$ are required to go zero -- like the length of D3 in the bridge 
example discussed earlier. One could introduce a parameter $\alpha$, such that the parameters $\alpha \uth^0$ give a continuous set of intermediate structures between 
$S_t$ and $S^*$, where $\alpha = 1$ implies $S_t$ and $\alpha = 0$ implies $S^*$. Now, suppose that there is a further set of parameters $\uth^1$ that must be deformed
into a set $\uth^2$ for $S^*$. In this case, one simply parametrises by a $\beta$, such that the intermediate structures have $\beta \uth^1 + (1-\beta) \uth_2$. 

\begin{figure}[htbp!]
    \centering
    \includegraphics[width=0.4\textwidth]{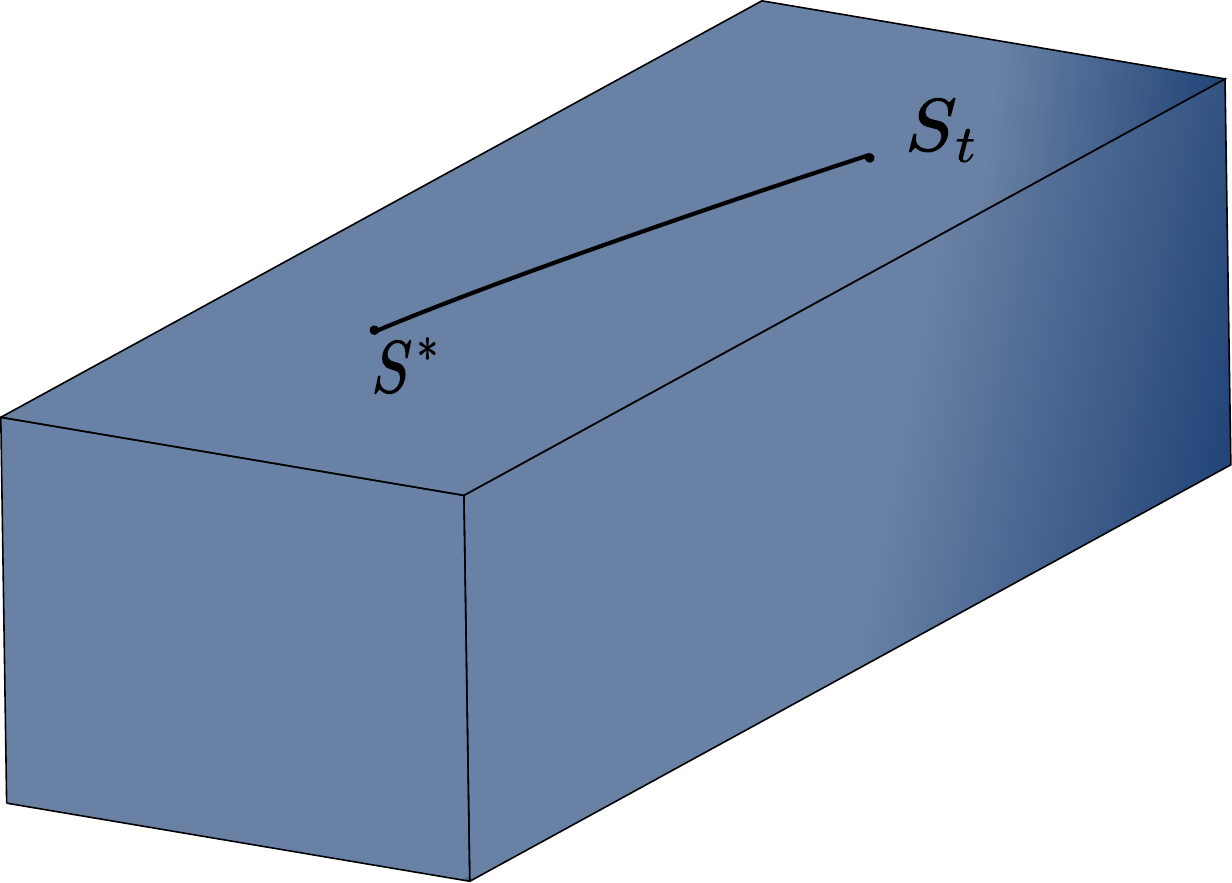}
    \caption{Transfer between a target structure and interpolating structure within a parametric family.}
    \label{fig:tf_ti}
\end{figure}

These specifications are important, because, if two structures $S_t$ and $S^*$ are in the same parametric family then they will have the same AG topology and the metric
distance between them will only {\em depend} on the attributes. In that case, one could simply use Euclidean distance as the metric on the space of parameters. Now, as the
cuboids corresponding to the parametric families are convex, locally in the $S_t$, one would have a straight line or {\em geodesic} between $S_t$ and $S^*$, parametrised
by $\alpha$ and $\beta$ (Figure \ref{fig:tf_ti}).

This condition means that, if one needed to break the transfer into more, smaller, steps, any intermediate structures between $S_t$ and $S^*$ can be taken on the geodesic,
thus automatically minimising the distance travelled between transfers.

\begin{figure}[htbp!]
    \centering
    \includegraphics[width=0.5\textwidth]{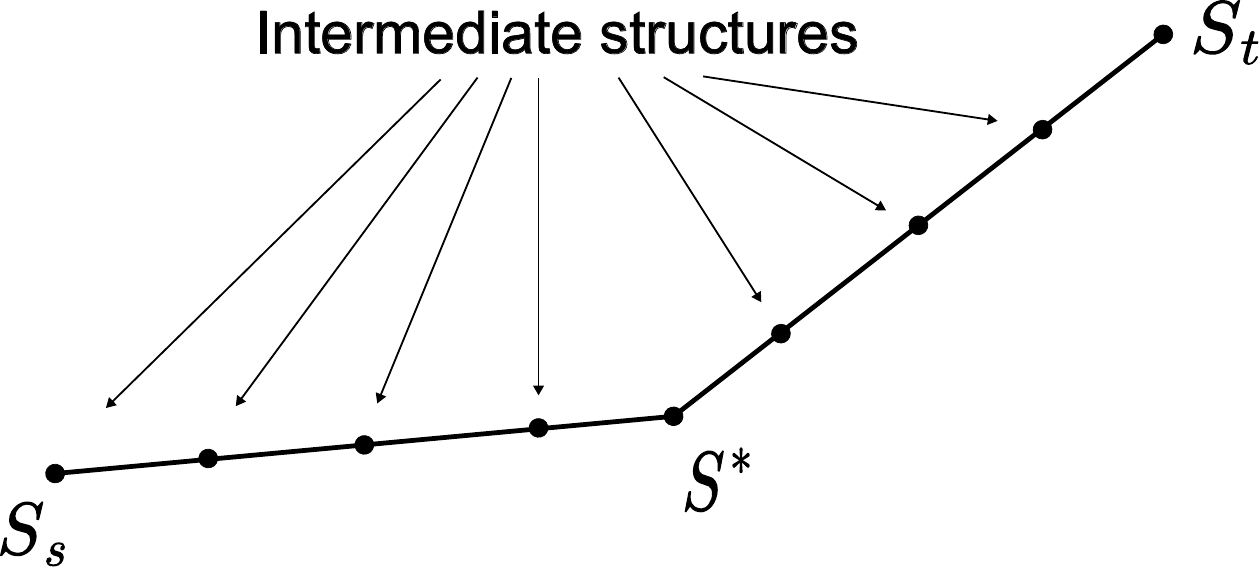}
    \caption{Transfer between two parametric families in terms of interpolating structures arranged on two geodeesics.}
    \label{fig:tf_2geo}
\end{figure}

Furthermore, as $S^*$ has a representation within the parametric family of $S_s$, one can parametrise the path between $S^*$ and $S_s$ with an $\alpha$ and $\beta$ and get
another geodesic (Figure \ref{fig:tf_2geo}).

The question now is how to choose an $S^*$ in such a way as to minimise $D(S_s,S^*) + D(S^*,S_t)$; the problem being that the two metrics are actually different and may need
to be weighted differently in the optimisation. The simplest way out is if, as previously discussed, one identifies a common parametric family large enough to encompass the 
families of $S_s$ and $S_t$. In this case, one can simply adopt the Euclidean metric in the big parameter space and then $S^*$ is simply identified as the mid point of the 
straight line (geodesic) between $S_t$ and $S_s$. In this case, where might geometrically-motivated transfer arise? One possibility is that the curve between $S_s$ and $S_t$
in the base space, will lift into a curve in the total space and could be used to motivate a `flow-based' algorithm for transfer; something analogous to parallel transport
\cite{Kobayashi} might be possible.

It is important to think in terms of curves rather than lines because features will not necessarily depend linearly on base-space parameters; i.e., natural frequencies 
$f \approx \sqrt{m} \approx \sqrt{\rho}$ -- straight lines in parameter space my not lift to straight lines between feature spaces.

\begin{figure}[htbp!]
    \centering
    \includegraphics[width=0.6\textwidth]{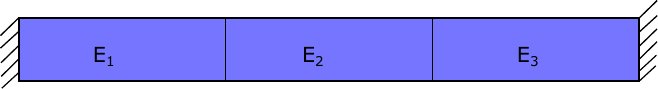}
    \caption{Schematic of simple structure illustrating damage localisation issue.}
    \label{fig:tf_dlo}
\end{figure}

Another important consideration at this point is that another form of compatibility might be needed, related to the SHM problems under consideration. Suppose that the root
problem is one of {\em damage localisation}. In this case, the obvious strategy is to allocate labels to the IEs in which one needs to find damage. Consider the cartoon 
system shown in Figure \ref{fig:tf_dlo}.

\begin{figure}[htbp!]
    \centering
    \includegraphics[width=0.35\textwidth]{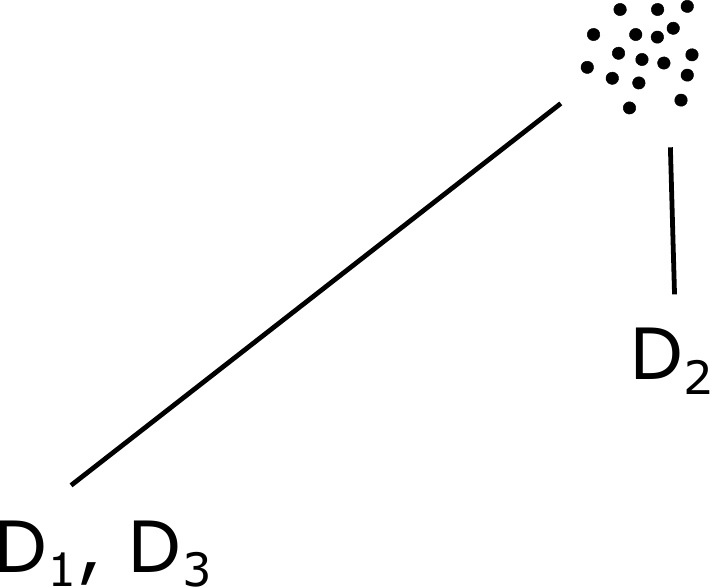}
    \caption{Illustration of feature data distributions for damage localisation problem.}
    \label{fig:fd_dlo}
\end{figure}

One could assign labels 1-3 to elements E1 to E3 in order to train a classifier. Suppose the feature vector in this case is the first two natural frequencies. When damage
occurs, one might see the situation illustrated in Figure \ref{fig:fd_dlo}.

\begin{figure}[htbp!]
    \centering
    \includegraphics[width=0.5\textwidth]{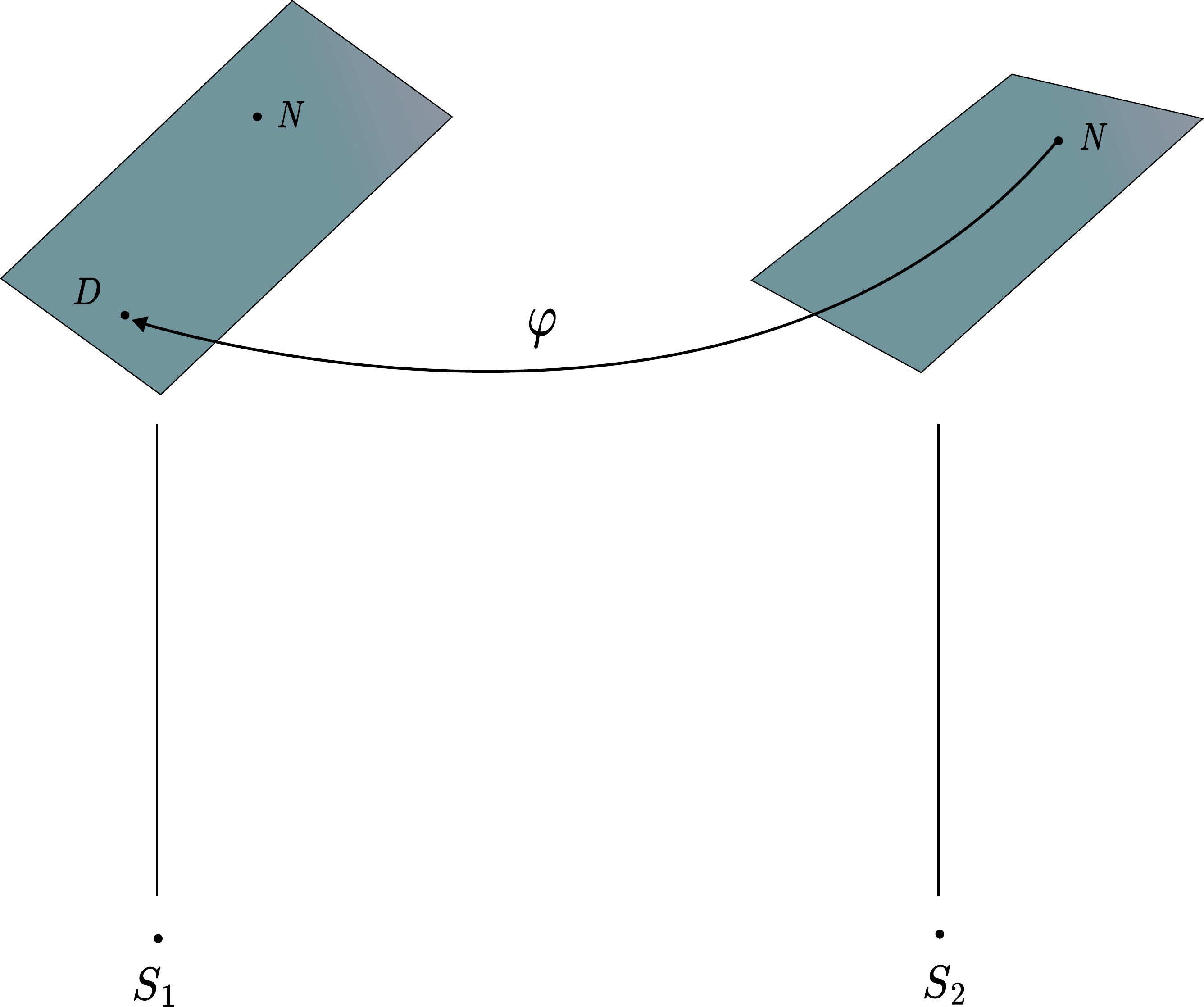}
    \caption{Illustration of the idea that neighbouring points in a parametric family could be both the healthy and damaged states of a single structure or the healthy 
             states of two different structures.}
    \label{fig:hdhh}
\end{figure}

Now, if damage were to manifest as, or could be represented by, a bulk change in IE properties, one could potentially represent D1 as a reduction of Young's modulus on E1.
However, it is important to note that, as this is just a change within the parametric family, the damaged structure -- say $S_1$ could coincide with a neighbouring healthy
structure -- say $S_2$. If one has a geometric mechanism for transfer, one might expect that $\phi$ from the normal condition of $S_2$ maps to D1 for S1 (Figure \ref{fig:hdhh}).

This consideration also needs to fit with any normalisation of the feature data. For example, if one were to use {\em Normal Condition Alignment} \cite{Poole} in all the
fibres to facillitate transfer, this would remove the mean of the normal condition data. In geometrical terms, this action would put all features for undamaged structures 
at zero, so the normal condition data would be on the zero section of the bundle throughout \cite{PBSHM4}.

The idea of combining data from models and real structures is an old one in SHM; in fact it is the basis of both the FE-updating approach \cite{Friswell} and so-called 
{\em forward-model driven} SHM \cite{Wilson}. In fact, PBSHM generalises both these approaches; in terms of FE-updating, the model need not be so close that the model's and structure's 
data be identified, just close enough that transfer is positive.

At this point it is worth mentioning that the idea of using intermediate structures is not fanciful. The idea of this paper has been to present an underpinning language,
actual transfer via intermediates has already been presented sucessfully in \cite{Dardeno1}.

\section{Conclusions}
\label{sec:conc}

This paper has looked at the geometry of population-based SHM, based on a fibre-bundle picture proposed in a previous paper. It makes two main contributions. In the first
place, the geometry of the feature spaces -- which are identified with the fibres in the picture -- is discussed. This part of the paper is largely empty in mathematical 
terms and is mainly intended to simply define terminology and notation which will unambiguously allow discussion of the various data operations which will be needed in the 
computational framework of PBSHM. The second contribution is to extend the purely formal analogy with fibre bundles to a practical contribution to the theory; this is 
accomplished by providing a topology for the `space of structures' which is the base space of the bundle. This result is made possible by defining parametric families of structures,
which allow a definition of open sets in the space and thus a notion of continuity; this then leads to the idea of {\em interpolating structures}, which may well provide a
key technology for transfer learning across heterogeneous populations of structures.  

A follow-on paper will look at the possibilities for geometric transfer even if the space of structures retains a discrete structure.

\section*{Acknowledgements}

The authors of this paper would like to thank Dr Chandula Wickramarachchi for providing the schematic used in Figure \ref{fig:fb_scheme}. The authors also gratefully acknowledge the 
support of the UK Engineering and Physical Sciences Research Council (EPSRC) via grant reference EP/W005816/1. For the purpose of  open access, the author has applied a Creative Commons 
Attribution (CC BY) licence to any Author Accepted Manuscript version arising.

\bibliography{isma_24_KW_1}

\begin{thebibliography}{10}
\providecommand{\url}[1]{#1}
\csname url@samestyle\endcsname
\providecommand{\newblock}{\relax}
\providecommand{\bibinfo}[2]{#2}
\providecommand{\BIBentrySTDinterwordspacing}{\spaceskip=0pt\relax}
\providecommand{\BIBentryALTinterwordstretchfactor}{4}
\providecommand{\BIBentryALTinterwordspacing}{\spaceskip=\fontdimen2\font plus
\BIBentryALTinterwordstretchfactor\fontdimen3\font minus \fontdimen4\font\relax}
\providecommand{\BIBforeignlanguage}[2]{{%
\expandafter\ifx\csname l@#1\endcsname\relax
\typeout{** WARNING: IEEEtran.bst: No hyphenation pattern has been}%
\typeout{** loaded for the language `#1'. Using the pattern for}%
\typeout{** the default language instead.}%
\else
\language=\csname l@#1\endcsname
\fi
#2}}
\providecommand{\BIBdecl}{\relax}
\BIBdecl

\bibitem{Farrar}
C.~Farrar and K.~Worden, \emph{Structural Health Monitoring: a Machine Learning Perspective}.\hskip 1em plus 0.5em minus 0.4em\relax John Wiley \& Sons, 2012.

\bibitem{WordenP}
K.~Worden, L.~Bull, P.~Gardner, J.~Gosliga, T.~Rogers, E.~Cross, E.~Papatheou, W.~Lin, and N.~Dervilis, ``An overview of recent developments in population-based structural health monitoring,'' \emph{Frontiers in the Built Environment}, vol.~6, p. Art.146, 2020.

\bibitem{PBSHM4}
G.~Tsialiamanis, C.~Mylonas, E.~Chatzi, N.~Dervilis, D.~Wagg, and K.~Worden, ``Foundations of population-based {SHM}, {P}art {IV}: Geometry of spaces of structures and their feature spaces,'' \emph{Mechanical Systems and Signal Processing}, vol. 157, p. 107692, 2021.

\bibitem{Kobayashi}
S.~Kobayashi and K.~Nomizu, \emph{Foundations of Differential Geometry}.\hskip 1em plus 0.5em minus 0.4em\relax Wiley, 2009.

\bibitem{PBSHM3}
P.~Gardner, L.~Bull, J.~Gosliga, N.~Dervilis, and K.~Worden, ``Foundations of population-based {SHM}, {P}art {III}: Heterogeneous populations -- transfer and mapping,'' \emph{Mechanical Systems and Signal Processing}, vol. 149, p. 107142, 2021.

\bibitem{Worden}
K.~Worden, P.~Gardner, E.~Cross, R.~Barthorpe, and D.~Wagg, ``On digital twins, mirrors and virtualisations,'' \emph{Journal of Risk and Uncertainty Part B: Mechanical Engineering}, vol.~6, p. 030902, 2020.

\bibitem{Peeters}
B.~Peeters and G.~D. Roeck, ``One-year monitoring of the {Z24}-{B}ridge: environmental effects versus damage events,'' \emph{Earthquake Engineering and Structural Dynamics}, vol.~30, pp. 149--171, 2001.

\bibitem{PanB}
Q.~Yang, Y.~Zhang, W.~Dai, and S.~Pan, \emph{Transfer Learning}.\hskip 1em plus 0.5em minus 0.4em\relax Cambridge University Press, 2020.

\bibitem{Pan}
S.~Pan and Q.~Yang, ``A survey on transfer learning,'' \emph{IEEE Transactions on Knowledge and Data Engineering}, vol.~22, pp. 1345--1359, 2010.

\bibitem{Pan2}
S.~Pan, I.~Tsang, J.~Kwok, and Q.~Yang, ``Domain adaptation via transfer component analysis,'' \emph{IEEE Transactions on Neural Networks}, vol.~22, pp. 199--210, 2011.

\bibitem{Poole}
J.~Poole, P.~Gardner, N.~Dervilis, L.~Bull, and K.~Worden, ``On statistic alignment for domain adaptation in population-based structural health monitoring,'' \emph{Structural Health Monitoring -- An International Journal}, vol.~22, pp. 1581--1600, 2023.

\bibitem{Friswell}
M.~Friswell and J.~Mottershead, ``Inverse methods in structural health monitoring,'' \emph{Key Engineering Materials}, vol. 204, pp. 201--210, 2001.

\bibitem{Wilson}
J.~Wilson, G.~Manson, P.~Gardner, and R.~Barthorpe, ``Hierarchical verification and validation in a forward model-driven structural health monitoring strategy,'' \emph{Structural Health Monitoring -- An International Journal}, p. 14759217231206698, 2023.

\bibitem{Dardeno1}
T.~Dardeno, L.~Bull, N.~Dervilis, and K.~Worden, ``Transfer learning via intermediate structures,'' in \emph{European Workshop on Structural Health Monitoring, Potsdam, Germany}, 2024.

\end{thebibliography}

\end{document}